\documentstyle[latex-acl,named]{article}
\def\psappears#1{
}

\psappears{Appears in ACL '94}

\begin{document}

\author{David R. Traum \and James F. Allen\\
Department~of Computer Science \\ University of Rochester \\ Rochester,
NY 14627-0226 \\ {\tt traum@cs.rochester.edu} \and {\tt james@cs.rochester.edu}
}

\title{Discourse Obligations in Dialogue Processing}

\maketitle

\begin{abstract}
We show that in modeling social interaction, particularly dialogue,
the attitude of {\em obligation} can be a useful adjunct to the
popularly considered attitudes of belief, goal, and intention and
their mutual and shared counterparts. In particular, we show how {\em
discourse obligations} can be used to account in a natural manner for
the connection between a question and its answer in dialogue and how
obligations can be used along with other parts of the discourse
context to extend the coverage of a dialogue system.\\
\end{abstract}

\pagestyle{plain}
\section{Motivation}

Most computational models of discourse are based primarily on an
analysis of the intentions of the speakers (e.g.,
\cite{Cohen79,Allen80,GroszSidner86}).  An agent has certain goals, and
communication results from a planning process to achieve these goals.
The speaker will form intentions based on the goals and then act on
these intentions, producing utterances.  The hearer will then
reconstruct a model of the speaker's intentions upon hearing the
utterance.  This approach has many strong points, but does not provide
a very satisfactory account of the adherence to discourse conventions
in dialogue.

For instance, consider one simple phenomena: a question is typically
followed by an answer, or some explicit statement of an inability or
refusal to answer. The intentional story account of this goes as
follows. From the production of a question by Agent B, Agent A
recognizes Agent B's goal to find out the answer, and she adopts a
goal to tell B the answer in order to be co-operative.  A then plans
to achieve the goal, thereby generating the answer.  This provides an
elegant account in the simple case, but requires a strong assumption
of co-operativeness.  Agent A must adopt agent B's goals as her own.
As a result, it does not explain why A says anything when she does not
know the answer or when she is not predisposed to adopting B's goals.

        Several approaches have been suggested to account for this
behavior.  \cite{Litman87} introduced an intentional analysis at the
discourse level in addition to the domain level, and assumed a set of
conventional multi-agent actions at the discourse level. Others have
tried to account for this kind of behavior using social intentional
constructs such as {\em Joint intentions} \cite{Cohen91a} or {\em
Shared Plans} \cite{Grosz90a}. While these accounts do help explain
some discourse phenomena more satisfactorily, they still require a
strong degree of co-operativity to account for dialogue coherence, and
do not provide easy explanations of why an agent might act in cases
that do not support high-level mutual goals.

Consider a stranger approaching an agent and asking, ``Do you have the
time?''  It is unlikely that there is a joint intention or shared
plan, as they have never met before. From a purely strategic point of
view, the agent may have no interest in whether the stranger's goals
are met. Yet, typically agents will still respond in such situations.

As another example, consider a case in which the agent's goals are
such that it prefers that an interrogating agent not find out the
requested information. This might block the formation of an intention
to inform, but what is it that inspires the agent to respond at all?

As these examples illustrate, an account of question answering must go
beyond recognition of speaker intentions. Questions do more than just
provide evidence of a speaker's goals, and something more than
adoption of the goals of an interlocutor is involved in formulating a
response to a question.

Some researchers, e.g., \cite{Mann88,Kowtko91}, assume a library of
discourse level actions, sometimes called dialogue games, which encode
common communicative interactions. To be co-operative, an agent must
always be participating in one of these games. So if a question is
asked, only a fixed number of activities, namely those introduced by a
question, are co-operative responses.  Games provide a better
explanation of coherence, but still require the agents to recognize
each other's intentions to perform the dialogue game. As a result,
this work can be viewed as a special case of the intentional view. An
interesting model is described by \cite{Airenti93}, which separates
out the conversational games from the task-related games in a way
similar way to \cite{Litman87}. Because of this separation, they do
not have to assume co-operation on the tasks each agent is performing,
but still require recognition of intention and co-operation at the
conversational level.  It is left unexplained what goals motivate
conversational co-operation.

The problem with systems which impose co-operativity in the form of
automatic goal adoption is that this makes it impossible to reason
about cases in which one might want to violate these rules, especially
in cases where the conversational co-operation might conflict with the
agent's personal goals.

      We are developing an alternate approach that takes a step back
from the strong plan-based approach.  By the strong plan-based
account, we mean models where there is a set of personal goals which
directly motivates all the behavior of the agent.  While many of the
intuitions underlying these approaches seems close
to right, we claim it is a mistake to attempt to analyze this
behavior as arising entirely from the agent's high-level goals.

	We believe that people have a much more complex set of
motivations for action. In particular, much of one's behavior arises
from a sense of obligation to behave within limits set by the society
that the agent is part of.  A model based on obligations differs from
an intention-based approach in that obligations are independent of
shared plans and intention recognition. Rather, obligations are the
result of rules by which an agent lives.  Social interactions are
enabled by their being a sufficient compatibility between the rules
affecting the interacting agents.  One responds to a question because
this is a social behavior that is strongly encouraged as one grows up,
and becomes instilled in the agent.

\section{Sketch of Solution}

        The model we propose is that an agent's behavior is determined
by a number of factors, including that agent's current goals in the
domain, and a set of obligations that are induced by a set of social
conventions. When planning, an agent considers both its goals and
obligations in order to determine an action that addresses both to the
extent possible. When prior intentions and obligations conflict, an
agent generally will delay pursuit of its intentions in order to
satisfy the obligations, although the agent may behave otherwise at
the cost of violating its obligations. At any given time, an agent may
have many obligations and many different goals, and planning involves
a complex tradeoff between these different factors.

        Returning to the example about questions, when an agent is
asked a question, this creates an obligation to respond. The agent
does not have to adopt the goal of answering the question as one of
her personal goals in order to explain the behavior. Rather it is a
constraint on the actions that the agent may plan to do. In fact, the
agent might have an explicit goal not to answer the question, yet
still is obliged to offer a response (e.g., consider most politicians
at press conferences). The planning task then is to satisfy the
obligation of responding to the question, without revealing the answer
if at all possible. In cases where the agent does not know the answer,
the obligation to respond may be discharged by some explicit
statement of her inability to give the answer.

\section{Obligations and Discourse Obligations}

Obligations represent what an agent {\em should} do, according to some
set of norms. The notion of obligation has been studied for many
centuries, and its formal aspects are examined using Deontic Logic.
Our needs are fairly simple, and do not require an extensive survey of
the complexities that arise in that literature. Still, the intuitions
underlying that work will help to clarify what an obligation is.
Generally, obligation is defined in terms of a modal operator often
called {\em permissible}. An action is {\em obligatory} if it is not
permissible not to do it. An action is {\em forbidden} if it is not
permissible. An informal semantics of the operator can be given by
positing a set of rules of behavior R. An action is obligatory if its
occurrence logically follows from R, and forbidden if its
non-occurrence logically follows from R. An action that might occur or
not-occur according to R is neither obligatory nor forbidden.

        Just because an action is obligatory with respect to a set of
rules R does not mean that the agent will perform the action. So we do
not adopt the model suggested by \cite{ShoTen92} in which agents' behavior
cannot violate the defined social laws.  If an obligation is not
satisfied, then this means that one of the rules must have been
broken. We assume that agents generally plan their actions to violate
as few rules as possible, and so obligated actions will usually
occur. But when they directly conflict with the agent's personal
goals, the agent may choose to violate them.  Obligations are quite
different from and can not be reduced to intentions and goals. In
particular, an agent may be obliged to do an action that is contrary
to his goals (for example, consider a child who has to apologize for
hitting her younger brother).

Obligations also cannot be reduced to simple {\em expectations},
although obligations may act as a source of expectations.  Expectations
can be used to guide the action interpretation and plan-recognition
processes (as proposed by \cite{Carberry90}), but expectations do not in
and of themselves provide a sufficient motivation for an agent to
perform the expected action -- in many cases there is nothing wrong
with doing the unexpected or not performing an expected action. The
interpretation of an utterance will often be clear even without
coherence with prior expectations. We need to allow for the
possibility that an agent has performed an action even when this
violates expectations. If an agent actually violates obligations as
well then the agent can be held accountable.\footnote{\cite{McRoy93}
uses expectations derived from Adjacency Pair structure
\cite{Schegloff73}, as are many of the discourse obligations
considered in this paper. These expectations correspond to social
norms and do impose the same notion of accountability. However, the
analysis there is oriented towards discovering misconceptions based on
violated expectations, and the alternative possibility of violated
obligations is not considered in the utterance recognition process,
nor allowed in the utterance production process.}

Specific obligations arise from a variety of sources. In a
conversational setting, an accepted offer or a promise will incur an
obligation.  Also, a command or request by the other party will bring
about an obligation to perform the requested action.  If the
obligation is to say something then we call this a {\em discourse
obligation}.  Our model of obligation is very simple. We use a set of
rules that encode discourse conventions. Whenever a new conversation
act is determined to have been performed, then any future action that
can be inferred from the conventional rules becomes an obligation.  We
use a simple forward chaining technique to introduce obligations.

Some obligation rules based on the performance of conversation acts
are summarized in Table~\ref{obl-table}.  When an agent performs a
promise to perform an action, or performs an acceptance of a
suggestion or request by another agent to perform an action, the agent
obliges itself to achieve the action in question. When another agent
requests that some action be performed, the request itself brings an
obligation to address the request: that is, either to accept it or to
reject it (and make the decision known to the requester) -- the
requestee is not {\em permitted} to ignore the request. A question
establishes an obligation to answer the question. If an utterance has
not been understood, or is believed to be deficient in some way, this
brings about an obligation to repair the utterance.

\begin{table}[ht]
\begin{tabular}{ll}
{\bf source of obligation} & {\bf obliged action}\\ \hline \hline
$\rm S_{1}$ Accept or Promise A & $\rm S_{1}$ achieve A\\\hline
$\rm S_{1}$ Request A& $\rm S_{2}$ address Request:\\
& \hspace*{0.125in} accept A {\bf or} reject A\\\hline
$\rm S_{1}$ YNQ whether P & $\rm S_{2}$ Answer-if P \\\hline
$\rm S_{1}$ WHQ P(x) & $\rm S_{2}$ Inform-ref x \\\hline
utterance not understood & repair utterance\\
\hspace*{0.125in}or incorrect &\\\hline
%$\rm S_{1}$ Init & $\rm S_{2}$ acknowledge\\\hline
%Request Repair of P & Repair P\\\hline
%Request Acknowledgement of P & acknowledge P\\\hline
\end{tabular}
\caption{\label{obl-table} Sample Obligation Rules}
\end{table}

\subsection{Obligations and Behavior}
Obligations (or at least beliefs that the agent has obligations) will
thus form an important part of the reasoning process of a deliberative
agent, e.g., the architecture proposed by
\cite{BratmanEtAl88}. In addition to considering {\bf beliefs} about the
world, which will govern the {\em possibility} of performing actions
and likelyhood of success, and {\bf desires} or goals which will
govern the {\em utility} or desirability of actions, a social agent
will also have to consider {\bf obligations}, which govern the {\em
permissibility} of actions.

There are a large number of strategies which may be used to incorporate
obligations into the deliberative process, based on how much weight
they are given compared to the agents goals. \cite{Conte93} present
several strategies of moving from obligations to actions, including:
automatically performing an obligated action, adopting all obligations
as goals, or adopting an obligated action as a goal only when
performing the action results in a state desired by the agent. In the
latter cases, these goals still might conflict with other goals of the
agent, and so are not guaranteed to be performed.

In general, we will want to allow action based on obligations to
supersede performance of intended actions. For instance, consider an
agent with an intention to do something as soon as possible. If an
obligation is imposed, it will still be possible to perform the
intended action, but a well-behaved agent might need to delay
performance until the obligation is dealt with. For example, if the
intention is to perform a series of inform acts, and then a listener
requests repair of one, a well-behaved agent will repair that inform
before proceeding to initiate the next intended one.

\section{Using Discourse Obligations in a Dialogue System} \label{trains-sec}

	We have built a system that explicitly uses discourse
obligations and communicative intentions to partake in natural
dialogue.  This system plays the role of the dialogue manager in the
TRAINS dialogue system, which acts as an intelligent planning
assistant in a transportation domain. While this is a domain where the
assumption of co-operation is generally valid, the obligation model
still provides for a much simpler analysis of the discourse behavior
than a strongly plan-based account.  An example of a dialogue that the
TRAINS system can engage in is shown in
Figure~\ref{warmup-dialog-fig}.
Below we describe parts of the discourse model in more detail and then
show how it is used to account for aspects of this dialogue.

\begin{figure}[ht]
\begin{tabular}{ll@{: }p{2.3in}}
{\bf Utt. \#} & \multicolumn{2}{l}{\bf Speaker: Utterance}\\
1   &    U & We better ship a boxcar of oranges to Bath by 8 AM. \\
2   &    S & Okay.\\
3-3=6 &  U & So we need to get a boxcar to Corning where there are oranges.\\
3-7  &   U & There are oranges at Corning.\\
3-8  &   U & Right?\\
4    &   S & Right. \\
5-1  &   U & So we need an engine to move the boxcar.\\
5-2  &   U & Right?\\
6    &   S & Right.\\
7-1=2  & U & So there's an engine at Avon.\\
7-3   &  U & Right?\\
8     &  S & Right. \\
9=13  &  U & So we should move the engine at Avon,  engine E1, to
Dansville to pick up the boxcar there.\\
14    &  S & Okay.  \\
15-2=4  & U & And move it from Dansville to Corning.\\
15-5=7  & U & Load up some oranges into the boxcar. \\
15-8=10 & U & And then move it on to Bath. \\
16     &    S & Okay.  \\
17     &   U & How does that sound?  \\
18-3  &     S  & That's no problem.    \\
19    &    U & Good.
\end{tabular}
\caption{\label{warmup-dialog-fig} Sample dialogue\footnotemark~ processed by
TRAINS-93.}
\end{figure}

\footnotetext{This is a slightly simplified
version of a spoken dialogue between two people. The original is
dialogue 91-6.1 from \cite{Gross-et-al92}. The utterance numbering
system used here reflects the relation to the turn and utterance
numbering used there.  `3-7' represents utterance 7 within turn 3. `='
is used to indicate merged utterances. Thus `3-3=6' spans four
utterances in turn 3 of the original, and 9=13 replaces turns 9 through
13 in the original.}

\begin{figure}[htb]
\setlength{\unitlength}{0.012500in}%
\begingroup\makeatletter\ifx\SetFigFont\undefined
% extract first six characters in \fmtname
\def\x#1#2#3#4#5#6#7\relax{\def\x{#1#2#3#4#5#6}}%
\expandafter\x\fmtname xxxxxx\relax \def\y{splain}%
\ifx\x\y   % LaTeX or SliTeX?
\gdef\SetFigFont#1#2#3{%
  \ifnum #1<17\tiny\else \ifnum #1<20\small\else
  \ifnum #1<24\normalsize\else \ifnum #1<29\large\else
  \ifnum #1<34\Large\else \ifnum #1<41\LARGE\else
     \huge\fi\fi\fi\fi\fi\fi
  \csname #3\endcsname}%
\else
\gdef\SetFigFont#1#2#3{\begingroup
  \count@#1\relax \ifnum 25<\count@\count@25\fi
  \def\x{\endgroup\@setsize\SetFigFont{#2pt}}%
  \expandafter\x
    \csname \romannumeral\the\count@ pt\expandafter\endcsname
    \csname @\romannumeral\the\count@ pt\endcsname
  \csname #3\endcsname}%
\fi
\fi\endgroup
\begin{picture}(192,299)(198,523)
\thinlines
\put(247,773){\vector( 0,-1){ 32}}
\put(342,725){\vector(-1, 0){ 47}}
\put(231,711){\vector( 0,-1){ 65}}
\put(279,630){\line( 1, 0){ 63}}
\put(342,630){\vector( 0, 1){ 49}}
\put(231,600){\vector( 0,-1){ 45}}
\put(342,711){\line( 0, 1){ 79}}
\put(342,790){\vector(-1, 0){ 42}}
\thicklines
\put(198,711){\framebox(97,30){}}
\put(310,679){\framebox(80,32){}}
\put(214,600){\framebox(65,46){}}
\put(231,779){\dashbox{4}(69,43){}}
\thinlines
\put(325,560){\line( 0, 1){ 50}}
\put(325,610){\vector(-1, 0){ 45}}
\thicklines
\put(215,523){\framebox(130,32){}}
\put(342,711){\makebox(0,0)[lb]{\smash{\SetFigFont{9}{10.8}{rm} }}}
\put(205,715){\makebox(0,0)[lb]{\smash{\SetFigFont{10}{12.0}{rm}
Modules}}}
\put(198,725){\makebox(0,0)[lb]{\smash{\SetFigFont{10}{12.0}{rm}  NL
Interpretation}}}
\put(231,679){\makebox(0,0)[lb]{\smash{\SetFigFont{10}{12.0}{rm} Observed}}}
\put(342,646){\makebox(0,0)[lb]{\smash{\SetFigFont{10}{12.0}{rm}  Intended}}}
\put(342,633){\makebox(0,0)[lb]{\smash{\SetFigFont{10}{12.0}{rm}  Conversation
Acts}}}
\put(225,610){\makebox(0,0)[lb]{\smash{\SetFigFont{10}{12.0}{rm}  Manager}}}
\put(235,580){\makebox(0,0)[lb]{\smash{\SetFigFont{10}{12.0}{rm} Domain
Directives}}}
\put(342,725){\makebox(0,0)[lb]{\smash{\SetFigFont{10}{12.0}{rm}  NL Output}}}
\put(247,757){\makebox(0,0)[lb]{\smash{\SetFigFont{10}{12.0}{rm} NL Input}}}
\put(225,625){\makebox(0,0)[lb]{\smash{\SetFigFont{10}{12.0}{rm}  Dialogue}}}
\put(240,525){\makebox(0,0)[lb]{\smash{\SetFigFont{10}{12.0}{rm}  Modules}}}
\put(220,540){\makebox(0,0)[lb]{\smash{\SetFigFont{10}{12.0}{rm}  Domain Task
Interaction}}}
\put(335,595){\makebox(0,0)[lb]{\smash{\SetFigFont{10}{12.0}{rm} Domain
Observations}}}
\put(255,795){\makebox(0,0)[lb]{\smash{\SetFigFont{10}{12.0}{bf}User}}}
\put(230,665){\makebox(0,0)[lb]{\smash{\SetFigFont{10}{12.0}{rm} Conversation
Acts}}}
\put(335,580){\makebox(0,0)[lb]{\smash{\SetFigFont{10}{12.0}{rm} and Directive
Responses}}}
\put(310,685){\makebox(0,0)[lb]{\smash{\SetFigFont{10}{12.0}{rm}
Module}}}
\put(315,695){\makebox(0,0)[lb]{\smash{\SetFigFont{10}{12.0}{rm}  NL
Generation}}}
\end{picture}
\caption{Dialogue Manager's High-Level View of the Architecture of the
TRAINS Conversation System \label{dm-arch1}}
\end{figure}
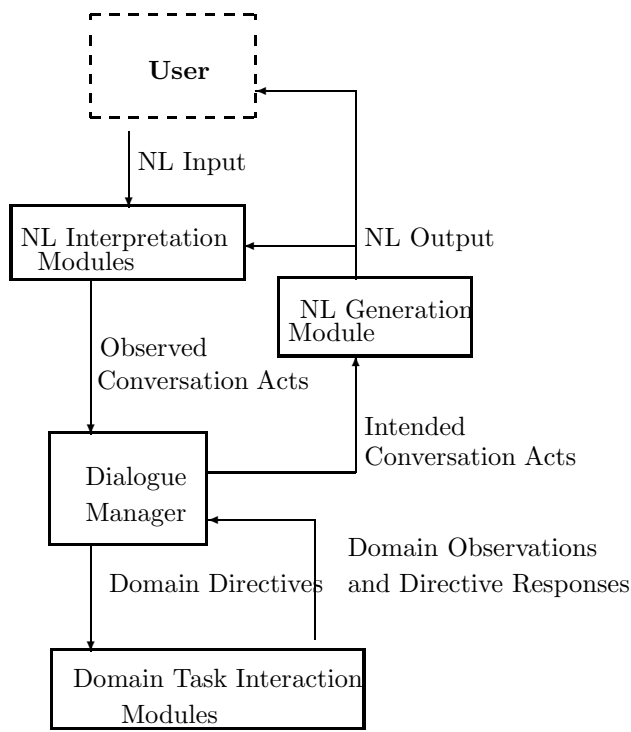

The TRAINS System \cite{Allen91} is a large integrated natural
language conversation and plan reasoning system. We concentrate here,
however, on just one part of that system, the discourse actor which
drives the actions of the dialogue manager module.
Figure~\ref{dm-arch1} illustrates the system from the viewpoint of the
dialogue manager.

The dialogue manager is responsible for maintaining the flow of
conversation and making sure that the conversational goals are met.
For this system, the main goals are that an executable plan which
meets the user's goals is constructed and agreed upon by both the
system and the user and then that the plan is executed.

The dialogue manager must keep track of the current state of the
dialogue, determine the effects of observed conversation acts,
generate utterances back, and send commands to the domain plan
reasoner and domain plan executor when appropriate. Conversational
action is represented using the theory of {\em Conversation Acts}
\cite{Traum92a} which augments traditional {\em Core Speech Acts} with
levels of acts for turn-taking, grounding \cite{Clark89}, and
argumentation. Each utterance will generally contain acts (or partial
acts) at each of these levels.

\subsection{Representing Mental Attitudes}
As well as representing general obligations within the temporal logic
used to represent general knowledge, the system also maintains two
stacks (one for each conversant) of pending {\bf discourse
obligations}. Each obligation on the stack is represented as an
obligation type paired with a content. The stack structure is
appropriate because, in general, one must respond to the most recently
imposed obligation first. As explained in Section~\ref{actor-sec}, the
system will attend to obligations before considering other parts of
the discourse context. Most obligations will result in the formation
of {\em intentions} to communicate something back to the user.  When
the intentions are formed, the obligations are removed from the stack,
although they have not yet actually been met. If, for some reason, the
system dropped the intention without satisfying it and the obligation
were still current, the system would place them back on the stack.

The over-riding goal for the TRAINS domain is to construct and execute
a plan that is shared between the two participants. This leads to
other goals such as accepting proposals that the other agent has
suggested, performing domain plan synthesis, proposing to the other
agent plans that the domain plan reasoner has constructed, or
executing a completed plan.

\subsection{The Discourse Actor Algorithm} \label{actor-sec}

In designing an agent to control the behavior of the dialogue manager,
we choose a {\em reactive} approach in which the system will not
deliberate and add new intentions until after it has performed the
actions which are already intended. As shown above, though, new
obligations will need to be addressed before performing intended
actions. The agent's deliberative behavior could thus be characterized
in an abstract sense as:

%\begin{figure}
\begin{tabbing}
12\=1234123\=\kill
{\bf loop}\\
\> perceive world and update beliefs\\
\>{\bf if} \>system has obligations\\
\>{\bf then} \>address obligations\\
\>{\bf else if} \>system has performable intentions\\
\>{\bf then} \> perform actions\\
\>{\bf else } \>deliberate on goals
\end{tabbing}
%\end{figure}

When deciding what to do next, the agent first considers obligations
and decides how to update the intentional structure (add new goals or
intentions) based on these obligations. Obligations might also lead
directly to immediate action. If there are no obligations, then the
agent will consider its intentions and perform any actions which it
can to satisfy these intentions. If there are no performable
intentions, then the system will deliberate on its overall goals and
perhaps adopt some new intentions (which can then be performed on the
next iteration).

For the discourse actor, special consideration must be given to the
extra constraints that participation in a conversation imposes. This
includes some weak general obligations (such as acknowledging
utterances by others and not interrupting) as well as some extra goals
coming from the domain setting to maintain a shared view of the world
and the domain plans which are to be executed. We prioritize the
sources for the deliberations of the actor as follows:\\
%\begin{enumerate}{}{\itemsep 0in\parsep 0in}
%\item
\hspace*{0.125in}1. Discourse Obligations from Table~\ref{obl-table}\\
%\item
\hspace*{0.125in}2. Weak Obligation: Don't interrupt user's turn\\
%\item
\hspace*{0.125in}3. Intended Speech Acts\\
%\item
\hspace*{0.125in}4. Weak Obl: Grounding (coordinate mutual beliefs)\\
%\item
\hspace*{0.125in}5. Discourse Goals: Domain Plan Negotiation \\
%\item
\hspace*{0.125in}6. High-level Discourse Goals\\
%\end{enumerate}
The implemented actor serializes consideration of these sources into
the algorithm in Figure~\ref{actor-alg}.

\begin{figure}[htbp]
\small
\begin{tabbing}
1234\=123\=12345\=123\=123\=123\=\kill
(1)\>{\bf while} conversation is not finished\\
(2)\>\>{\bf if} system has obligations\\
(3)\>\> {\bf then} address obligations\\
(4)\>\> {\bf else if} system has turn\\
(5)\>\> {\bf then if} system has intended conversation acts\\
(6)\>\>\> {\bf then} call generator to produce NL utterances\\
(7)\>\>\> {\bf else if} some material is ungrounded\\
(8)\>\>\>{\bf then} address grounding situation\\
(9)\>\>\> {\bf else if}  some proposal is not accepted\\
(10)\>\>\>{\bf then} consider proposals\\
(11)\>\>\> {\bf else if} high-level goals are unsatisfied\\
(12)\>\>\>{\bf then} address goals\\
(13)\>\>\> {\bf else} release turn\\
\>\>\>\> or attempt to end conversation\\
(14)\>\>{\bf else if} no one has turn\\
(15)\>\> {\bf then} take turn\\
(16)\> \>{\bf else if} long pause\\
(17)\>\> {\bf then} take turn
\end{tabbing}
\caption{Discourse Actor Algorithm \label{actor-alg}}
\end{figure}

The updating of the conversational state due to perceived conversation
acts or actions of other modules of the system progresses
asynchronously with the operation of the discourse actor.  Whenever
the discourse actor is active, it will first decide on which task to
attempt, according to the priorities given in Figure~\ref{actor-alg},
and then work on that task. After completing a particular task, it
will then run through the loop again, searching for the next task,
although by then the context may have changed due to, e.g., the
observance of a new utterance from the user. The actor is always
running and decides at each iteration whether to speak or not
(according to turn-taking conventions); the system does not need to
wait until a user utterance is observed to invoke the actor, and need
not respond to user utterances in an utterance by utterance fashion.

Lines 2-3 of the algorithm in Figure~\ref{actor-alg} indicate that the
actor's first priority is fulfilling obligations. If there are any,
then the actor will do what it thinks best to meet those
obligations. If there is an obligation to address a request, the actor
will evaluate whether the request is reasonable, and if so, accept it,
otherwise reject it, or, if it does not have sufficient information to
decide, attempt to clarify the parameters.  In any case, part of
meeting the obligation will be to form an intention to tell the user
of the decision (e.g., the acceptance, rejection, or
clarification). When this intention is acted upon and the utterance
produced, the obligation will be discharged.  Other obligation types
are to repair an uninterpretable utterance or one in which the
presuppositions are violated, or to answer a question. In question
answering, the actor will query its beliefs and will answer depending
on the result, which might be that the system does not know the
answer.

In most cases, the actor will merely form the intention to produce the
appropriate utterance, waiting for a chance, according to turn-taking
conventions, to actually generate the utterance.  In certain cases,
though, such as a repair, the system will actually try to take control
of the turn and produce an utterance immediately.  For motivations
other than obligations, the system adopts a fairly ``relaxed''
conversational style; it does not try to take the turn until given it
by the user unless the user pauses long enough that the conversation
starts to lag (lines 14-17). When the system does not have the turn,
the conversational state will still be updated, but the actor will not
try to deliberate or act.

When the system does have the turn, the actor first (after checking
obligations) examines its intended conversation acts. If there are
any, it calls the generator to produce an utterance\footnote{Actually,
if the only utterance is an acknowledgement, the actor will postpone
the production until it checks that there is nothing else that it can
combine in the same utterance, such as an acceptance or answer.}
(lines 5-6 of the discourse actor algorithm). Whatever utterances are
produced are then reinterpreted (as indicated in
Figure~\ref{dm-arch1}) and the conversational state updated
accordingly.  This might, of course, end up in releasing the turn. It
might not be convenient to generate all the intended acts in one
utterance, in which case there will remain some intended acts left for
future utterances to take care of (unless the subsequent situation
merits dropping those intentions). Only intended speech acts that are
part of the same argumentation acts as those which are uttered will be
kept as intentions -- others will revert back to whatever caused the
intention to be formed, although subsequent deliberation might cause
the intentions to be re-adopted.

If there are no intended conversation acts, the next thing the actor
considers is the grounding situation (lines 7-8). The actor will try
to make it mutually believed (or {\em grounded}) whether particular
speech acts have been performed. This will involve acknowledging or
repairing user utterances, as well as repairing and requesting
acknowledgement of the system's own utterances. Generally, grounding
is considered less urgent than acting based on communicative
intentions, although some grounding acts will be performed on the
basis of obligations which arise while interpreting prior utterances.

If all accessible utterances are grounded, the actor then considers
the negotiation of domain beliefs and intentions (lines 9-10). The
actor will try to work towards a shared domain plan, adding intentions
to perform the appropriate speech acts to work towards this goal. This
includes accepting, rejecting, or requesting retraction of user
proposals, requesting acceptance of or retracting system proposals,
and initiating new system proposals or counterproposals.

The actor will first look for User proposals which are not shared. If
any of these are found, it will add an intention to accept the
proposal, unless the proposal is deficient in some way (e.g., it will
not help towards the goal or the system has already come up with a
better alternative). In this latter case, the system will reject the
user's proposal and present or argue for its own proposal. Next, the
actor will look to see if any of its own proposals have not been
accepted, requesting the user to accept them if they have been simply
acknowledged, or retracting or reformulating them if they have already
been rejected. Finally, the actor will check its private plans for any
parts of the plan which have not yet been proposed. If it finds any
here, it will adopt an intention to make a suggestion to the user.

If none of the more local conversational structure constraints
described above require attention, then the actor will concern itself
with its actual high-level goals. For the TRAINS system, this will
include making calls to the domain plan reasoner and domain executor,
which will often return material to update the system's private view of
the plan and initiate its own new proposals. It is also at this point
that the actor will take {\em control} of the conversation, pursuing
its own objectives rather than responding to those of the user.

Finally, if the system has no unmet goals that it can work towards
achieving (line 13), it will hand the turn back to the user or try to
end the conversation if it believes the user's goals have been met as
well.

\subsection{Examples}
The functioning of the actor can be illustrated by its behavior in the
dialogue in Figure~\ref{warmup-dialog-fig}.  While the discussion here
is informal and skips some details, the dialogue is actually processed
in this manner by the implemented system. More detail both on the
dialogue manager and its operation on this example can be found in
\cite{Traum94c}.

 Utterance 1 is interpreted as performing two Core Speech Acts. It is
interpreted (literally) as the initiation\footnote{According to the
theory of Conversation Acts \cite{Traum92a}, Core Speech Acts such as
inform are multi-agent actions which have as their effect a mutual
belief, and are not completed unless/until they are grounded.} of an
inform about an obligation to perform a domain action (shipping the
oranges). This utterance is also seen as (the initiation of) an
(indirect) suggestion that this action be the goal of a shared domain
plan to achieve the performance of the action. In addition, this
utterance releases the turn to the system.  Figure~\ref{ds-1} shows
the relevant parts of the discourse state after interpretation of this
utterance.

\begin{figure}[h]
\small
{\bf Discourse Obligations:}   \\
{\bf Turn Holder:}   System\\
{\bf Intended Speech Acts:}  \\
{\bf Unack'd Speech Acts: }  {\tt [INFORM-1], [SUGGEST-4]}\\
{\bf Unaccepted Proposals:}  \\
{\bf Discourse~Goals:}~{\tt Get-goal~Build-Plan~Execute-Plan}
\caption{\label{ds-1} Discourse Context after Utterance 1}
\end{figure}

After interpreting utterance 1, the system first decides to
acknowledge this utterance (lines 7-8 in the actor algorithm) --
moving the suggestion from an unacknowledged to unaccepted -- and then
to accept the proposal (lines 9-10). Finally, the system acts on the
intentions produced by these deliberations (lines 5-6) and produces
the combined acknowledgement/acceptance of utterance 2. This
acceptance makes the goal shared and also satisfies the first of the
discourse goals, that of getting the domain goal to work on.

Utterances 3-3=6 and 3-7 are interpreted, but not responded to yet
since the user keeps the turn (in this case by following up with
subsequent utterances before the system has a chance to
act). Utterance 3-8 invokes a discourse obligation on the system to
respond to the User's assertion in 3-7 and also gives the turn to the
system. The resulting discourse context (after the system decides to
acknowledge) is shown in Figure~\ref{ds-2}.

\begin{figure}[ht]
\small
{\bf Discourse Obligations:} {\tt (CHECK-IF (:AT ...))}\\
{\bf Turn Holder:}   System\\
{\bf Intended Speech Acts:}  {\tt (Ack [INFORM-7], ...)}\\
{\bf Unack'd Speech Acts:}  \\
{\bf Unaccepted Proposals:}  {\tt [SUGGEST-10], [SUGGEST-15]}\\
{\bf Discourse Goals:} {\tt Build-Plan Execute-Plan}
\caption{\label{ds-2} Discourse Context after Utterance 2}
\end{figure}

 The system queries its domain knowledge base and decides that the
user is correct here (there are, indeed, oranges at Corning), and so
decides to meet this obligation (lines 2-3) by answering in the
affirmative.  This results in forming an intention to inform, which is
then realized (along with the acknowledgement of the utterances) by
the production of utterance 4.

Similar considerations hold for the system responses 6 and 8. The
reasoning leading up to utterance 14 is similar to that leading to
utterance 2. Here the user is suggesting domain actions to help lead
to the goal, and the system, when it gets the turn, acknowledges and
accepts this suggestion.

Utterances 15-2=4, 15-5=7, and 15-8=10 are interpreted as requests
because of the imperative surface structure. The discourse obligation
to address the request is incurred only when the system decides to
acknowledge the utterances and ground them. After the decision to
acknowledge, the obligations are incurred, and the system then
addresses the requests, deciding to accept them all, and adding
intentions to perform an accept speech act, which is then produced as
16.

Utterance 17 is interpreted as a request for evaluation of the plan.
When the system decided to acknowledge, this creates a discourse
obligation to address the request.  The system considers this
(invoking the domain plan reasoner to search the plan for problems or
incomplete parts) and decides that the plan will work, and so decides
to perform the requested action -- an evaluation speech act. This is
then generated as 18-3. The discourse state after the decision to
acknowledge is shown in Figure~\ref{ds-3}.

\begin{figure}[ht]
\small
{\bf Discourse Obligations:} {\tt (ADDRESS [REQUEST-49])}\\
{\bf Turn Holder:}   System\\
{\bf Intended Speech Acts:}  {\tt (Ack [REQUEST-49])}\\
{\bf Unack'd Speech Acts:}  \\
{\bf Unaccepted Proposals:}  \\
{\bf Discourse Goals:} {\tt Build-Plan Execute-Plan}
\caption{\label{ds-3} Discourse Context after Utterance 17}
\end{figure}

After the user's assent, the system then checks its goals, and, having
already come up with a suitable plan, executes this plan in the
domain by sending the completed plan to the domain plan executor.

This example illustrates only a small fraction of the capabilities of
the dialogue model. In this dialogue, the system needed only to follow
the initiative of the user. However this architecture can handle
varying degrees of initiative, while remaining responsive. The default
behavior is to allow the user to maintain the initiative through the
plan construction phase of the dialogue.  If the user stops and asks
for help, or even just gives up the initiative rather than continuing
with further suggestions, the system will switch from plan recognition
to plan elaboration and incrementally devise a plan to satisfy the
goal (although this plan would probably not be quite the same as the
plan constructed in this dialogue).

We can illustrate the system behaving more on the basis of goals than
obligations with a modification of the previous example.  Here, the
user releases the turn back to the system after utterance 2, and the
deliberation proceeds as follows: the system has no obligations, no
communicative intentions, nothing is ungrounded, and there are no
unaccepted proposals, so the system starts on its high-level
goals. Given its goal to form a shared plan, and the fact that the
current plan (consisting of the single abstract {\tt move-commodity}
action) is not executable, the actor will call the domain plan
reasoner to elaborate the plan. This will return a list of
augmentations to the plan which can be safely assumed (including a
{\tt move-engine} event which {\em generates} the move-commodity,
given the conditions that the oranges are in a boxcar which is
attached to the engine), as well as some choice point where one of
several possibilities could be added (e.g., a choice of the particular
engine or boxcar to use).

Assuming that the user still has not taken the turn back, the system
can now propose these new items to the user. The choice could be
resolved in any of several ways: the domain executor could be queried
for a preference based on prior experience, or the system could put
the matter up to the user in the form of an alternative question, or
it could make an arbitrary choice and just suggest one to the user.

The user will now be expected to acknowledge and react to these
proposals. If the system does not get an acknowledgement, it will
request acknowledgement the next time it considers the grounding
situation. If the proposal is not accepted or rejected, the system can
request an acceptance. If a proposal is rejected, the system can
negotiate and offer a counterproposal or accept a counter proposal
from the user.

Since the domain plan reasoner \cite{Ferguson94b} performs both plan
recognition and plan elaboration in an incremental fashion, proposals
from system and user can be integrated naturally in a mixed-initiative
fashion. The termination condition will be a shared executable plan
which achieves the goal, and each next action in the collaborative
planning process will be based on local considerations.

\section{Discussion}

We have argued that obligations play an important role in accounting
for the interactions in dialog. Obligations do not replace the
plan-based model, but augment it. The resulting model more readily
accounts for discourse behavior in adversarial situations and other
situations where it is implausible that the agents adopt each others
goals.  The obligations encode learned social norms, and guide each
agent's behavior without the need for intention recognition or the use
of shared plans at the discourse level. While such complex intention
recognition may be required in some complex interactions, it is not
needed to handle the typical interactions of everyday discourse.
Furthermore, there is no requirement for mutually-agreed upon rules
that create obligations. Clearly, the more two agents agree on the
rules, the smoother the interaction becomes, and some rules are
clearly virtually universal.  But each agent has its own set of
individual rules, and we do not need to appeal to shared knowledge to
account for local discourse behavior.

We have also argued that an architecture that uses obligations
provides a much simpler implementation than the strong plan-based
approaches. In particular, much of local discourse behavior can arise
in a ``reactive manner'' without the need for complex planning. The
other side of the coin, however, is a new set of problems that arise
in planning actions that satisfy the multiple constraints that arise
from the agent's personal goals and perceived obligations.

The model presented here allows naturally for a mixed-initiative
conversation and varying levels of co-operativity. Following the
initiative of the other can be seen as an {\em obligation driven}
process, while leading the conversation will be {\em goal driven}.
Representing both obligations and goals explicitly allows the system
to naturally shift from one mode to the other. In a strongly
co-operative domain, such as TRAINS, the system can subordinate working
on its own goals to locally working on concerns of the user, without
necessarily having to have any shared discourse plan. In less
co-operative situations, the same architecture will allow a system to
still adhere to the conversational conventions, but respond in
different ways, perhaps rejecting proposals and refusing to answer
questions.

\section*{Acknowledgements}
This material is based upon work supported by ONR/DARPA under grant
number N00014-92-J-1512. We would like to thank the rest of the TRAINS
group at the University of Rochester for providing a stimulating
research environment and a context for implementing these ideas within
an integrated system.

\bibliographystyle{named}

\end{document}